\documentclass[a4paper,fleqn]{cas-dc}

\usepackage[numbers]{natbib}

\usepackage{pifont} 
\usepackage{xcolor, soul}
\sethlcolor{Apricot}

\def\tsc#1{\csdef{#1}{\textsc{\lowercase{#1}}\xspace}}
\tsc{WGM}
\tsc{QE}
\tsc{EP}
\tsc{PMS}
\tsc{BEC}
\tsc{DE}

\begin{document}
\let\WriteBookmarks\relax
\def\floatpagepagefraction{1}
\def\textpagefraction{.001}

\shorttitle{Localizing in-domain adaptation of biomedical language models}

\shortauthors{TM Buonocore et~al.}

\title [mode = title]{Localising In-Domain Adaptation of Transformer-Based Biomedical Language Models}                      


%
\author[1]{Tommaso Mario Buonocore}[orcid=0000-0002-2887-088X]
\cormark[1]
\ead{buonocore.tms@gmail.com}
\credit{Conceptualization, Methodology, Software, Investigation, Writing - Original Draft}

%
\author[2]{Claudio Crema}[orcid=0000-0003-2537-9742]
\credit{Conceptualization, Methodology, Software, Investigation}
\author[2]{Alberto Redolfi}[orcid=0000-0002-4145-9059]
\credit{Conceptualization, Supervision}
\author[1]{Riccardo Bellazzi}[orcid=0000-0002-6974-9808]
\credit{Supervision}
\author[1]{Enea Parimbelli}[orcid=0000-0003-0679-828X]
\credit{Conceptualization, Supervision, Writing - Review \& Editing}

\address[1]{Dept. of Electrical, Computer and Biomedical Engineering, University of Pavia,Pavia,27100, Italy}

\address[2]{Laboratory of Neuroinformatics, IRCCS Istituto Centro San Giovanni di Dio Fatebenefratelli,Brescia,25125,Italy}

\cortext[cor1]{Corresponding author}

\begin{abstract}
In the era of digital healthcare, the huge volumes of textual information generated every day in hospitals constitute an essential but underused asset that could be exploited with task-specific, fine-tuned biomedical language representation models, improving patient care and management. For such specialized domains, previous research has shown that fine-tuning models stemming from broad-coverage checkpoints can largely benefit additional training rounds over large-scale in-domain resources. However, these resources are often unreachable for less-resourced languages like Italian, preventing local medical institutions to employ in-domain adaptation. In order to reduce this gap, our work investigates two accessible approaches to derive biomedical language models in languages other than English, taking Italian as a concrete use-case: one based on neural machine translation of English resources, favoring quantity over quality; the other based on a high-grade, narrow-scoped corpus natively written in Italian, thus preferring quality over quantity. Our study shows that data quantity is a harder constraint than data quality for biomedical adaptation, but the concatenation of high-quality data can improve model performance even when dealing with relatively size-limited corpora. The models published from our investigations have the potential to unlock important research opportunities for Italian hospitals and academia. Finally, the set of lessons learned from the study constitutes valuable insights towards a solution to build biomedical language models that are generalizable to other less-resourced languages and different domain settings.
\end{abstract}


\maketitle

\section{Introduction}

The digitization of health services and clinical care processes has led healthcare organizations to routinely produce an ever-increasing number of textual data: medical reports, nursing notes, discharge letters, and insurance claims are just some of the types of digital documents clinicians must deal with daily \cite{wang_clinical_2018}. Due to its high informativeness, this source of information can be a key asset for medical applications assisted by artificial intelligence (AI), from biomedical text mining to clinical predictive modeling. The unstructured nature of textual data and the complexity of the biomedical domain have always been a challenge for AI developers, but the recent advancements in the field of natural language processing (NLP) brought by large-scale pretrained models based on the transformer architecture \cite{vaswani_attention_2017} offer new opportunities for advancing the state of the art of many biomedical-related tasks.

The remarkable success of transformer-based models like BERT (Bidirectional Encoder Representations from Transformers) \cite{devlin_bert_2019} and its derivatives \cite{clark_electra_2020, lan_albert_2020} is largely related to the pretrain-then-finetune paradigm used for training. The language model first undergoes a resource-demanding training procedure (i.e., pretraining) that uses a large volume of general-purpose textual data and extensive computational resources to learn the grammatical structures and the semantics of the language of interest. The pretraining phase is self-supervised, avoiding data labeling by adopting training objectives based on pseudo labels, like masked language modeling (MLM) \cite{devlin_bert_2019}. The pretrained model can then be adapted to serve different tasks after a second round of relatively inexpensive (and therefore accessible) training (i.e., fine-tuning) using labeled custom data that updates the weights of the original model according to the needs of the specific task and domain of application. Directly applying a general-purpose, pretrained model on biomedical problems, however, can underperform due to the prominent distributional differences between general domain texts and biomedical texts. 

When the target domain varies substantially from the pretraining corpus, as in biomedicine, models can benefit from an intermediate round of domain-adaptive training on large domain-specific corpora (e.g., biomedical literature, either abstracts or full text) with the same pretraining objectives \cite{lee_biobert_2020}. However, the availability of open-source, biomedical corpora sufficiently large to be used for domain adaptation is scarce due to the sensitive nature of health-related information, and essentially limited to the English language, in light of its established role as the language of science \cite{gordin_scientific_2015}. Same considerations can be drawn for knowledge bases as well, which have been proven to be useful to improve the performance of language models in downstream tasks \cite{xie_pre-trained_2022}, with non-English metathesauri covering only a minimal portion of their English counterpart (1.5\% for Italian UMLS, from 5 sources against 148 for English). These reasons lead biomedical domain adaptation of cutting-edge techniques like transformer-based language models to be often prohibitive for relatively small biomedical institutions and less-resourced languages.

\subsection{Objective} 

We want to investigate the less-resourced languages and thus we take Italian as our main use case, nonetheless drawing conclusions that are applicable to other languages that can be considered low-resourced in this context. Motivated by the limitations of available transformer-based solutions in the biomedical domain, as well as the lack of any publicly available biomedical corpora in Italian, this paper brings the following original contributions:
\begin{itemize}
  \item[A.] We reduce the gap between English and Italian biomedical NLP by developing a new biomedical checkpoint for the Italian language based on BioBERT, hereinafter referred to as BioBIT (Biomedical Bert for ITalian).
  \item[B.] We provide and evaluate an automated pipeline based on neural machine translation that can be applied to other less-resourced languages to overcome the difficulties of acquiring biomedical textual data in the target language.
  \item[C.] We investigate the effects of the pretraining data size and quality, evaluating them on manually curated in-domain sentences to better understand how biomedical knowledge is distilled into the models and to elucidate best practices regarding data requirements to obtain significant performance improvements.
\end{itemize}

The pretrained BioBIT model, the Italian pretraining corpora, and the source code are made publicly available\footnotemark{} as integral part of the publication, contributing to lowering the entry barrier to well-performing language models in less-resourced languages and medical domains.
\footnotetext{Code and data are available at https://github.com/IVN-RIN/bio-med-BIT, while the models can be found on the HuggingFace model repository at https://huggingface.co/IVN-RIN. Some  corpora cannot be shared for copyright reasons.}

\subsection{Related Work}
The widespread adoption of transformer-based models in many NLP tasks across different domains increased the need of domain-specific checkpoints (i.e., model snapshots stored in non-volatile memory), which boosted research on in-domain adaptation for language representation models. From a biomedical perspective, the first and most well-known pretrained model is BioBERT \cite{lee_biobert_2020}, a biomedical language representation model that shares the same architecture with BERT. Following the domain-adaptation approach, BioBERT is initialized using BERT weights that were pretrained on general domain texts; these weights are then updated using biomedical pretraining corpora, outperforming the former model and achieving state-of-the-art results in a variety of biomedical text mining tasks like clinical concept recognition, gene-protein relation extraction or biomedical question answering. In order to collect enough open-source biomedical data, Lee et al. leveraged biomedical literature repositories like PubMed and PMC, acquiring 4.5B words from abstracts and 13.5B words from full-text articles. A similar approach is followed by SciBERT, which uses the original BERT configuration but replaces the initial general domain corpora with 1.14M scientific articles randomly selected from Semantic Scholar. This corpus is composed by 82\% broad biomedical domain papers and 18\% papers from the computer science domain. By training from scratch on biomedical data, SciBERT can use a custom dictionary to better reflect the in-domain word distribution. These two strategies have later been updated either in terms of model architecture, replacing BERT with its variants \cite{naseem_bioalbert_2021,ozyurt_effectiveness_2020}, or in terms of in-domain pretraining data, extending the biomedical corpus based on scientific literature with other sources \cite{alsentzer_publicly_2019, chakraborty_biomedbert_2020}.

This wide variety of biomedical BERT-based models is favored by the wide availability of publicly accessible biomedical data in English, like MIMIC \cite{johnson_mimic-iii_2016}, the largest open-access dataset of medical records, and vast repositories of biomedical scientific literature \cite{national_institutes_of_health_national_nodate}. Aside from English, the majority of languages lack access to these valuable resources, which makes it hard to meet the expectations set by their English equivalent. Nevertheless, researchers from different countries attempted to pretrain non-English biomedical checkpoints, leveraging local (and often not publicly available) biomedical text collections, either training a new model from scratch \cite{akhtyamova_named_2020} or applying biomedical domain adaptation over multilingual \cite{schneider_biobertpt_2020} or monolingual \cite{copara_contextualized_2020} versions of BERT. 

For what concerns Italian, to the best of our knowledge, no such research effort has been described in the literature, which motivated us further in pursuing this work.

\begin{figure*}
    \centering
    \includegraphics[width=.9\linewidth]{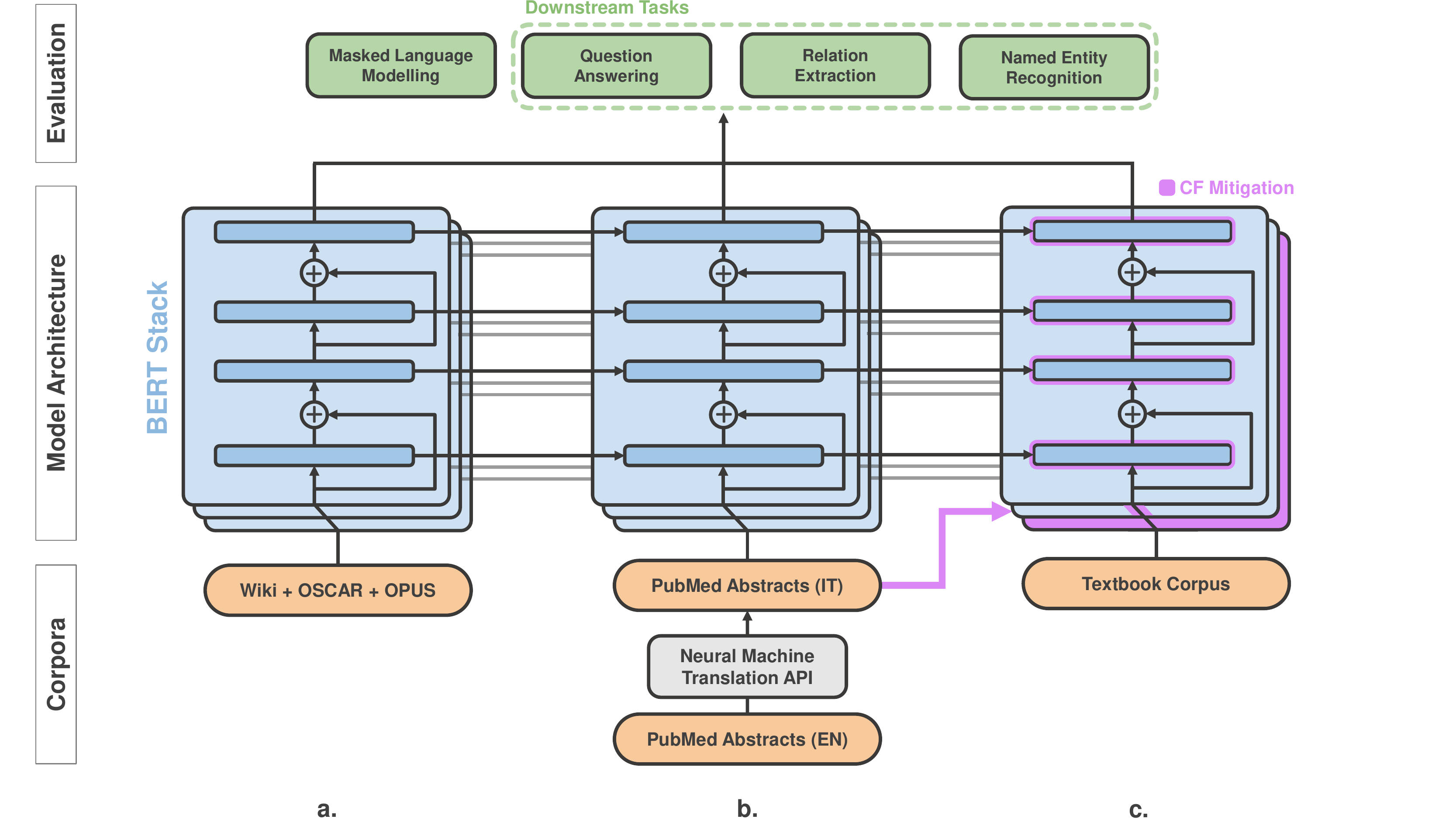}
    \caption{Training and evaluation pipeline for the BaseBIT model (a.), pretrained on various Italian corpora of generic text and used as the baseline, the BioBIT (b.) model, derived from machine-translated biomedical abstracts, and the MedBIT (c.) model, obtained using a corpus of selected medical texts natively written in Italian.}
    \label{fig:overallpic}
\end{figure*}

\section{Methods}

\subsection{Experimental Setting}
The overall process is illustrated in Figure \ref{fig:overallpic} and can be described as follows: starting from a general-purpose, Italian checkpoint (a.), we derived several biomedical adaptations following three main strategies: prioritizing the size of the training set (quantity over quality, b.); prioritizing the quality of the data, although limited in size (quality over quantity, c.); concatenating the two strategies (quality after quantity, b. and c.). Each model has been evaluated on MLM and the best-performing models on a battery of popular biomedical downstream tasks. The details of each step of the pipeline are presented in the following sub-sections. 

\subsection{Italian Language Pretraining}
To pretrain BioBIT, we followed the general approach outlined in BioBERT, built on the foundation of the BERT architecture. The pretraining objective is a combination of MLM (also known as the Cloze task) and next sentence prediction (NSP). The MLM objective is based on randomly masking 15\% of the input sequence, trying then to predict the missing tokens; for the NSP objective, instead, the model is given a couple of sentences and has to guess if the second comes after the first in the original document.

Our model has been initialized with a monolingual Italian version of BERT\footnotemark{}, obtained from a recent Wikipedia dump and various texts in Italian from the OPUS and OSCAR corpora collection, which sums up to the final corpus size of 81GB and 13B tokens. We used this checkpoint, simply referred to as BaseBIT, as the baseline. At the time of designing our work, no other large-scale, pretrained BERT or BERT-derived checkpoints were available for Italian. \footnotetext{Model repository: https://huggingface.co/dbmdz/bert-BaseBITalian-xxl-cased.}

At the time of writing, over 12 thousand BERT-based models (\~8\% of the total) are hosted in the Huggingface model repository, covering more than 20 different non-English languages. The unmatched popularity of BERT in the NLP community makes it the best candidate for our study, which will be easily replicable in different non-English-speaking countries.

\subsection{Quantity over quality: machine-translated PubMed abstracts}
Due to the unavailability of an Italian equivalent for the millions of abstracts and full-text scientific papers used by English, BERT-based biomedical models, in this work we leveraged machine translation to obtain an Italian biomedical corpus based on PubMed abstracts and train BioBIT. For this purpose, we adopted Google’s neural machine translation (NMT) system \cite{wu_googles_2016}, a framework that uses a combination of transformers and recurrent neural networks (RNNs) to achieve accurate translation for over 100 languages \cite{shen_lingvo_2019}, Italian included \cite{aiken_updated_2019}. While not as good as human translation, NMT has also been shown to work well in clinical settings, such as for translating abstractions of clinical trials published in languages other than English \cite{jackson_accuracy_2019}. 

The novelty of our approach from what concerns NMT is not in the engine itself, which we take as it is, but in the way it is employed. It is common practice for less-resourced languages to leverage translation (usually relying upon time-consuming manual revision) in the opposite direction, to unlock the opportunity offered by the many biomedical NLP tools available only for English \cite{becker_extraction_2016}. In our study, instead, we investigate whether NMT systems are mature enough to do the opposite, starting from the English source (in our case, the large biomedical corpus made of PubMed articles, which are solely written in English and therefore have no local equivalent) to develop local tools (e.g., clinical concept taggers) without any supervision.

To keep the in-domain model compatible with the general-purpose model, BaseBIT and BioBIT share the same vocabulary. Thanks to the WordPiece tokenization, any out-of-vocabulary biomedical word can still be dealt with.

\subsection{Quality over quantity: Italian medical textbooks}
Albeit biomedical-specific, PubMed abstracts remain a relatively heterogeneous textual resource, concerning a broad spectrum of subdomains ranging from the characterization of metal nanoparticles in mollusks to the evaluation of clinical practice guidelines for patient mobilization. This diversity, coupled with the unavoidable degradation introduced by machine translation, led us to formulate the hypothesis that our model might benefit from an additional round of more narrow-scoped, high-quality, strictly medical data natively written in Italian. 

To test this hypothesis, we collected a corpus of medical textbooks, either directly written by Italian authors or translated by human professional translators, used in formal medical doctors' education and specialized training. The size of such corpus amounts to 100 MB of data, corresponding to 0.35\% the size of the PubMed corpus used in \cite{lee_biobert_2020}. Given their educational nature, we believe that these comprehensive collections, if sufficiently large, of medical concepts can impact the encoding of biomedical knowledge in language models, with the advantage of being natively available in a wide variety of languages and not only English. Models trained on the textbook corpus are referred to as MedBIT. Online healthcare information dissemination is another source of biomedical texts that is commonly available in many less-resourced languages. Therefore, we also gathered an additional 100 MB of web-crawled data from reliable Italian, health-related websites, augmenting the size of our quality-prioritized corpus to 0.70\% of the PubMed corpus. The MedBIT models trained on the augmented corpus are marked as MedBIT\textsuperscript{+}.

\subsection{Catastrophic Forgetting Mitigation}
Subsequent pretraining of deep models on small corpora, like the textbook one, is known to be prone to Catastrophic Forgetting (CF), which translates into degradation of performance of the further-trained model compared to the pretrained baseline \cite{mccloskey_catastrophic_1989}. In other words, when catastrophic forgetting happens the network tends to fit the new input data distribution (i.e., the medical setting provided by medical textbooks) interfering with previously acquired knowledge (i.e., the broader biomedical setting provided by translated PubMed abstracts). To prevent the model parameters learned on new data from significantly deviating from previously learned parameters, during the training of MedBIT we tested combinations of different techniques originating from Continual Learning \cite{kirkpatrick_overcoming_2017}, approaching catastrophic forgetting mitigation either in terms of learning regularization (Layer-wise Learning Rate Decay \cite{zhang_revisiting_2021}, Warmup \cite{mccloskey_catastrophic_1989}, Layer Freezing) or knowledge distillation (Mixout \cite{lee_mixout_2020}, Experience Replay \cite{de_masson_d_autume_episodic_2019}, \cite{lin_self-improving_1992}). The best models enhanced with catastrophic forgetting mitigation have been labeled MedBIT\textsubscript{R} when presenting and discussing results in the following sections.

Layer-wise learning rate decay (LLRD) applies a layer-wise decay function to the learning rate so that layers closer to the input nodes, which often encode more common, general, and broad-based information, will have a smaller learning rate than the layers closer to the output that encodes localized information. Optionally, the learning rate schedule can be initialized with a short warm-up (WU) phase. Following the same rationale of LLRD,  layer freezing (LF) sets the gradient of the deepest layers of the network to zero before the last training phase, so that the weights encoding the more general information remain unaltered. The mixout (M) approach, instead, stochastically mixes the weights of the pretrained checkpoint and the ones of the model currently under training, which improves the stability of language model tuning even for a small number of training examples. 

While mixout regularization leverages only the parameters of the pretrained model, experience replay (ER) samples observations from a replay buffer. In the context of language modeling, we implemented experience replay by feeding the model with an additional batch of random data coming from the previous pretrained checkpoint every $n$ steps, where $n$ is a tunable hyperparameter called replay frequency. Note that, in contrast to previously mentioned approaches, ER requires to have access to not only the pretrained checkpoint but also the pretraining corpus itself, which is not the common case in pretrained models.

\subsection{Language Modeling Evaluation}
To better investigate the effects of data quality and quantity in pretraining our biomedical Italian models, we created different checkpoints pretrained on different sizes of the same corpus, named Model\textsubscript{XS/S/M/L}, as reported in Table \ref{table:1}. To make results comparable, corpora are incremental, i.e., bigger collections are obtained by appending additional text to the previous one. The MLM performance for pretraining has been evaluated using the average pseudo-perplexity (PPPL) metric \cite{salazar_masked_2020}, defined as: 

\begin{equation}
\label{eqn:pppl}
\operatorname{PPPL}(C):=\exp \left(-\frac{1}{N} \sum_{S \in C} \sum_{t=1}^{|S|} \log p\left(w_t \mid S_{\backslash t}\right)\right)
\end{equation}

where $N$ denotes the number of tokens in the corpus \(C\), \(w_t\) is $t$-th word of the sentence S of the corpus, and \(p(w_t|S_{\backslash t})\) indicates the conditional probability of the model for the masked word given the other words of the sentence. In order to specifically evaluate the progression of biomedical knowledge encoding during pretraining, we also checked the top five tokens predicted by the model on a set of manually-curated sentences, assigning a score according to the ranking \(R_{{w_M},S}\)  of the correct word $w_M$ in the top five tokens $T_5$ proposed by the model for each masked sentence. The Mean Reciprocal Rank (MRR) obtained this way is therefore simply defined as in Equation \ref{eqn:mrr}.

\begin{equation}
\label{eqn:mrr}
\operatorname{MRR}(C):=\frac{1}{|C|} \sum_{S \in C} \frac{1}{R_{w_M{ }^S}} \text { if } w_M \subset T_5, 0 \text { otherwise }
\end{equation}

The dataset used for internal MLM evaluation stems from 54 sentences in Italian pertaining to the medical domain, collected manually from reliable online resources like medical associations (e.g., AIRC\footnotemark{}\footnotetext{https://www.iss.it/}), institutions (e.g., ISS\footnotemark{}\footnotetext{https://www.airc.it/}), or journals (e.g., Italian Journal of Cardiology\footnotemark{}\footnotetext{https://www.giornaledicardiologia.it/}). For each sentence, we manually masked only the words that were medical-related and effectively reasonable to guess from the context provided (i.e., the rest of the sentence), resulting in 897 maskings. The sentences have different lengths, spanning from 39 to 258 tokens, and come from different sub-domains. The sub-domain distributions of the sentences for MLM evaluation and the textbook corpus have been compared in Figure A2 in Supplementary Notes.

\subsection{Downstream Task Evaluation}
Even though pretraining evaluation is the main focus of our work, we acknowledge that what matters the most when delivering a new checkpoint to the community is how the fine-tuned models initialized with it perform on target downstream tasks. Therefore, we fine-tuned our pretrained models on the three common biomedical tasks, namely named entity recognition (NER), extractive question answering (QA), and relation extraction (RE), relying on the same datasets previous work has been evaluated on. All the sources are in English, and no Italian equivalents are available at the time of writing. Therefore, we adapted them to the Italian language through neural machine translation, adopting the same strategy used for the BioBIT biomedical corpus. For extractive QA, we translated the benchmark datasets 4b, 5b and 6b of the BioAsq challenge \cite{tsatsaronis_overview_2015}. For NER, we selected six heterogeneous datasets (BC5CDR \cite{li_biocreative_2016}, BC2GM \cite{smith_overview_2008}, NCBI \cite{dogan_ncbi_2014}, Species-800 \cite{pafilis_species_2013}, BC4CHEMD \cite{krallinger_chemdner_2015}) annotating chemicals, diseases, genes, and other biomedical-related mentions. For RE, we evaluated the CHEMPROT \cite{taboureau_chemprot_2011} translated dataset, which annotates chemical-protein interactions, and the BioRED \cite{luo_biored_2022} dataset, which includes multiple entity types (e.g. disease, chemical, gene) and relation pairs (e.g. gene-disease; chemical–chemical) at the document level. Datasets have been translated using the same automatic translation procedure described for PubMed abstracts, correcting the start and end indices for each annotation after the translation. Comparisons between pre-trained corpus and downstream corpora are reported in Table A4, A5, and Figure A1 in Supplementary Notes. We kept the pre-assigned train/dev/test splits where available, autonomously splitting the datasets randomly otherwise. The fine-tuning procedure has been repeated five times for each model, initializing each run with a different random state. 

\section{Results}

\subsection*{First Experiment: BioBIT}
In the first experiment, we test the feasibility of training an Italian biomedical checkpoint, relying upon the machine-translated version of the PubMed abstracts used to train the original BioBERT in English. With reference to Figure \ref{fig:overallpic}, we therefore compare the baseline BaseBIT model (a.) with the BioBIT model (b.). The aim is to cover a common limitation for many local institutions, i.e., the lack of large and publicly available biomedical corpora for less-resourced languages. Even in presence of the appropriate  computational power, the lack of adequately sized input data can be a major barrier for language modeling. This experiment, therefore, checks if modern neural machine translation is sufficiently good to overcome this limitation, following the idea that the spurious patterns introduced by the translation process can be considered neglectable compared with the useful ones on a large scale. The results of our first experiment are shown in Table \ref{table:1}, along with the details of the corpora used in the pretraining of each model.

\begin{table}[width=\linewidth,cols=7,pos=h]
\caption{Details about corpora and metrics used for the pretraining evaluation of each model. The BaseBIT corpus is flagged as partially translated because it is made of multiple corpora, and some of them come from translated sources. Corpus size is expressed in terms of gigabytes.}\label{table:1}
\begin{tabular*}{\tblwidth}{@{} LCCCCCC@{} }
\toprule
\multirow{2}{*}{Model} & \multicolumn{4}{c}{Corpus} & \multicolumn{2}{c}{MLM Score} \\
\cmidrule{2-5}\cmidrule{6-7}
 & Size & Pretrain & Domain & Translated &  MRR & PPPL \\
\midrule
\begin{tabular}[c]{@{}l@{}}\textbf{BaseBIT}\\ \textbf{(baseline)}\end{tabular} & 81 & n.a. & General & Partial & 0.343 & \textbf{2.374}  \\
BioBIT\textsubscript{XS} & 0.1 & BaseBIT & Biomed & Yes & 0.343 & 2.453  \\
BioBIT\textsubscript{S} & 0.3 & BaseBIT & Biomed & Yes & 0.354 & 2.592  \\
BioBIT\textsubscript{M} & 1 & BaseBIT & Biomed & Yes & 0.352 & 2.837  \\
\textbf{BioBIT\textsubscript{L}} & 28 & BaseBIT & Biomed & Yes & \textbf{0.383} & 3.350  \\
\bottomrule
\end{tabular*}
\end{table}

\subsection*{Second Experiment: MedBIT}
In the second experiment, instead of relying on large machine-translated biomedical corpora, we pretrain on a small-sized corpus made of medical textbooks originally written in Italian, as described in Methods. This experiment aims to check if having high-quality, narrow-scoped data is enough to overcome the need for large-scale datasets for pretraining. The pretraining is done both starting from the original BaseBIT checkpoint (MedBIT\textsubscript{OR}) or concatenated after the BioBIT pretraining (MedBIT). For the concatenated MedBIT model, we tested pretraining both on the regular textbook corpora and the version augmented with web-crawled data (MedBIT\textsuperscript{+}). Results are shown in Table \ref{table:2}, with details about the catastrophic forgetting mitigation configuration where applicable (MedBIT\textsubscript{R}). Table \ref{table:2} showcases the best-performing configurations for CF mitigation with different techniques, while a comprehensive comparison of all the CF-mitigation configurations tested is available in Supplementary Notes.

\begin{table}[width=\linewidth,cols=9,pos=h]
\caption{Pretraining evaluation of each MedBIT model. R\textsubscript{n} = trained using CF mitigation techniques in different configurations. R\textsuperscript{+} = trained as in R, but using the augmented textbook corpus. O = trained skipping the BioBIT checkpoint.}
\label{table:2}
\begin{tabular*}{\tblwidth}{@{} LCCCCCCCC@{} }
\toprule
\multirow{2}{*}{Model} & \multirow{2}{*}{Pretrain} & \multicolumn{5}{c}{CF Mitigation} & \multicolumn{2}{c}{MLM Score} \\
\cmidrule{3-7}\cmidrule{8-9}
& & LLRD & LF & ER & M & WU & MRR & PPPL \\
\midrule
\begin{tabular}[c]{@{}l@{}}BaseBIT\\ (baseline)\end{tabular} & n.a. & \ding{55} & \ding{55} & \ding{55} & \ding{55} & \ding{55} & 0.343  & 2.374 \\
MedBIT\textsubscript{OR} & BaseBIT   & 0.9  & \ding{55}  & \ding{55}  & \ding{51}  & 0.02 & 0.365  & 2.203 \\
MedBIT & BioBIT\textsubscript{L} & \ding{55} & \ding{55} & \ding{55} & \ding{55} & \ding{55} & 0.365  & 2.389 \\
MedBIT\textsubscript{RF} & BioBIT\textsubscript{L} & \ding{55} & 6    & \ding{55} & \ding{55} & \ding{55} & 0.370  & 2.403 \\
MedBIT\textsubscript{R0} & BioBIT\textsubscript{L} & 0.9  & \ding{55} & 100  & \ding{55} & \ding{55} & 0.376  & 2.253 \\
MedBIT\textsubscript{R3} & BioBIT\textsubscript{L} & 0.9  & \ding{55} & \ding{55} & 0.9  & 0.02 & 0.375  & 2.279 \\
MedBIT\textsubscript{R3}\textsuperscript{+} & BioBIT\textsubscript{L} & 0.95 & \ding{55} & \ding{55} & 0.9  & 0.02 & 0.378  & 2.168 \\
\textbf{MedBIT\textsubscript{R12}\textsuperscript{+}} & BioBIT\textsubscript{L} & 0.95 & \ding{55} & 50   & \ding{55} & \ding{55} & \textbf{0.384} & \textbf{2.016} \\
\bottomrule
\end{tabular*}
\end{table}

\subsection*{Third Experiment: Fine-Tuning on Downstream Tasks}
In the last experiment, the baseline model (i.e., BaseBIT) and the best-performing models (i.e., BioBIT\textsubscript{L}, MedBIT\textsubscript{R12}\textsuperscript{+}, and its best non-ER alternative MedBIT\textsubscript{R3}\textsuperscript{+}) have been evaluated on a set of conventional biomedical downstream tasks. Due to the mismatches between exact answers, entities, and relations introduced by the automatic machine translation process, some examples have been dropped. For NER and RE, the number of mismatched examples is moderate, ranging from 0.3\% of the original size to 4.7\%. For QA, instead, the complexity of answers and contexts results in an 18-22\% drop. Details about translation-induced drops are reported in Supplementary Notes. For the sake of completeness, the multilingual BERT checkpoint (BERT\textsubscript{Multi}), compatible with Italian, has been evaluated as a second baseline for all the downstream tasks as well.
The results for each dataset of each downstream task in terms of F1 performance have been collected in Table \ref{table:3}, Table \ref{table:4}, and \ref{table:5}.

\begingroup
\setlength{\tabcolsep}{1.4pt}
\begin{table*}[cols=13,pos=h]
\caption{F1 performance of the fine-tuned models for NER on different biomedical datasets, reported in terms of mean and standard deviation over five different runs. \textsuperscript{†} Original dataset splits not provided. The \(\Delta \)\% reports the performance difference with the baseline.}
\label{table:3}
\begin{tabular}{@{}lcccccccccccc@{}}
\toprule
\multirow{3}{*}{Model} & \multicolumn{12}{c}{Dataset}\\ 
\cmidrule{2-13} & 
\multicolumn{2}{c}{BC2GM} & \multicolumn{2}{c}{BC4CHEMD} & 
\multicolumn{2}{c}{BC5CDR\textsubscript{CDR}} & \multicolumn{2}{c}{BC5CDR\textsubscript{DNER}} & 
\multicolumn{2}{c}{NCBI\_DISEASE} & \multicolumn{2}{c}{SPECIES-800} \\
\cmidrule(lr){2-3}\cmidrule(lr){4-5}\cmidrule(lr){6-7}\cmidrule(lr){8-9}\cmidrule(lr){10-11}\cmidrule(lr){12-13}
& Mean (sd)       & \(\Delta \)\%     & Mean (sd)        & \(\Delta \)\%       & Mean (sd)         & \(\Delta \)\%       & Mean (sd)  & \(\Delta \)\%        & Mean (sd)           & \(\Delta \)\%         & Mean (sd)         & \(\Delta \)\%         \\
\midrule
\begin{tabular}[c]{@{}l@{}}BaseBIT\\ (baseline)\end{tabular}    & 77.59 (.26)    & 0.0\%   & 76.43 (.20)     & 0.0\%     & 79.71 (.14)      & 0.0\%     & 69.68 (.46)      & 0.0\%      & 61.70 (1.09)        & 0.0\%       & 40.50 (2.24)      & 0.0\%       \\
 BERT\textsubscript{Multi} & 79.14 (.51)    & 2.0\%   & 77.53 (.43)     & 1.4\%     & 79.37 (.70)      & -0.4\%     & 72.86 (.45)      & 4.6\%      & 61.79 (1.85)        & 0.2\%       & 55.04 (1.52)      & 35.9\%       \\
\textbf{BioBIT\textsubscript{L}}                & \textbf{82.14 (.37)}    & \textbf{5.9\%}   & 80.70 (.16)     & 5.6\%     & 82.15 (.28)      & 3.1\%     & 76.27 (.42)      & 9.5\%      & \textbf{65.06 (1.23)}        & \textbf{5.5\%}       & 61.86 (.79)      & 52.7\%      \\
\textbf{MedBIT\textsubscript{R3}\textsuperscript{+}}              & 81.87 (.55)    & 5.4\%   & 80.68 (.29)     & 5.6\%     & 81.97 (.38)      & 2.8\%     & \textbf{76.32 (.35)}      & \textbf{9.5\%}      & 63.36 (.13)        & 2.7\%       & \textbf{63.90 (.58) }     & \textbf{57.8\%}      \\
\textbf{MedBIT\textsubscript{R12}\textsuperscript{+}   }           & 82.02 (.19)    & 5.7\%   & \textbf{80.75 (.22) }    & \textbf{5.7\%}     & \textbf{82.29 (.47) }     & \textbf{3.3\%}     & 75.65 (.98)      & 8.6\%      & 63.41 (.74)        & 2.8\%       & 63.02 (1.38)      & 55.6\%     \\
\bottomrule
\end{tabular}
\end{table*}
\endgroup

\begin{table*}[pos=h]
\caption{F1 performance of the fine-tuned models for QA on different biomedical datasets of factoid questions. The \(\Delta \)\% reports the performance difference with the baseline.}
\label{table:4}
\begin{tabular}{@{}lcccccc@{}}
\toprule
\multirow{3}{*}{Model} & \multicolumn{6}{c}{Datasets} \\
\cmidrule{2-7}
 & \multicolumn{2}{c}{BioASQ 4b} & \multicolumn{2}{c}{BioASQ 5b} & \multicolumn{2}{c}{BioASQ 6b} \\
 \cmidrule(lr){2-3}\cmidrule(lr){4-5}\cmidrule(lr){6-7}
 & Mean (sd) & \(\Delta \)\% & Mean (sd) & \(\Delta \)\% & Mean (sd) & \(\Delta \)\% \\
 \midrule
BaseBIT (baseline) & 68.38 (0.73) & 0.0\% & 77.69 (0.44) & 0.0\% & 73.83 (0.95) & 0.0\% \\
\textbf{BERT\textsubscript{Multi}}  & \textbf{71.28 (0.56)} & \textbf{4.2\%} & \textbf{79.27 (0.35)} & \textbf{2.0\%} & 75.15 (0.53) & 1.8\% \\
\textbf{BioBIT\textsubscript{L}} & 68.49 (0.71) & 0.2\% & 78.33 (0.56) & 0.8\% &\textbf{ 75.73 (0.71) }& \textbf{2.6\%} \\
MedBIT\textsubscript{R3}\textsuperscript{+} & 68.21 (0.62) & -0.3\% & 77.89 (0.46) & 0.3\% & 75.28 (0.35) & 2.0\% \\
MedBIT\textsubscript{R12}\textsuperscript{+} & 68.33 (0.93) & -0.1\% & 78.08 (0.15) & 0.5\% & 75.12 (1.24) & 1.8\% \\
\bottomrule
\end{tabular}
\end{table*}

\begin{table}[width=\linewidth,cols=5,pos=ht]
\caption{F1 performance of the fine-tuned models for RE on different biomedical datasets. The \(\Delta \)\% reports the performance difference with the baseline.}
\label{table:5}
\begin{tabular*}{\tblwidth}{@{} LCCCC@{}}
\toprule
\multirow{3}{*}{Model} & \multicolumn{4}{c}{Datasets} \\ 
\cmidrule{2-5}
 & \multicolumn{2}{c}{CHEMPROT} & \multicolumn{2}{c}{BioRED} \\
  \cmidrule(lr){2-3}\cmidrule(lr){4-5}
 & Mean (sd) & \(\Delta \)\% & Mean (sd) & \(\Delta \)\% \\
 \midrule
\begin{tabular}[c]{@{}l@{}}BaseBIT\\ (baseline)\end{tabular} & 34.88 (1.10) & 0.0\% & 63.15 (0.89) & 0.0\% \\ 
BERT\textsubscript{Multi} & 34.34 (0.59) & -1.7\% & 56.40 (2.00) & -10.7\% \\ 
BioBIT\textsubscript{L} & 38.16 (0.94) & 9.4\% & 67.15 (0.87) & 6.3\% \\ 
\textbf{MedBIT\textsubscript{R3}\textsuperscript{+}} & \textbf{38.82 (0.62)} & \textbf{11.3\%} & \textbf{67.62 (0.96)} & \textbf{7.1\%} \\
MedBIT\textsubscript{R12}\textsuperscript{+} & 37.37 (0.53) & 7.2\% & 67.37 (1.55) & 6.7\% \\ 
\bottomrule
\end{tabular*}
\end{table}

\section*{Discussion}
Allowing Italian healthcare institutions to access valuable biomedical checkpoints is an important step to unlock downstream research and medical applications leveraging unstructured clinical text, currently underused, through NLP. These checkpoints, however, commonly rely upon large-scale in-domain data that don’t have any Italian equivalent. This is a common limitation for less-resourced languages: we cannot manage to have data that are at the same time abundant and adequately narrow-scoped. This study explored the two main avenues to overcome this barrier, to assess which of them works better for the biomedical Italian setting, and to provide high-level insights that generalize to other less-resourced languages and application domains.

Our evaluation showed that the NMT-based Italian version of BioBERT, BioBIT\textsubscript{L}, outperformed the baseline model BaseBIT on any metric and task we have tested, either upstream (MLM evaluation through PPPL and MRR) or downstream (F1 performance on NER, QA and RE). This proves that neural machine translation can be leveraged to obtain localized versions of the English pretraining checkpoints for Italian when the size of the training corpus is big enough. 

\begin{figure}
\centering
\includegraphics[width=1\linewidth]{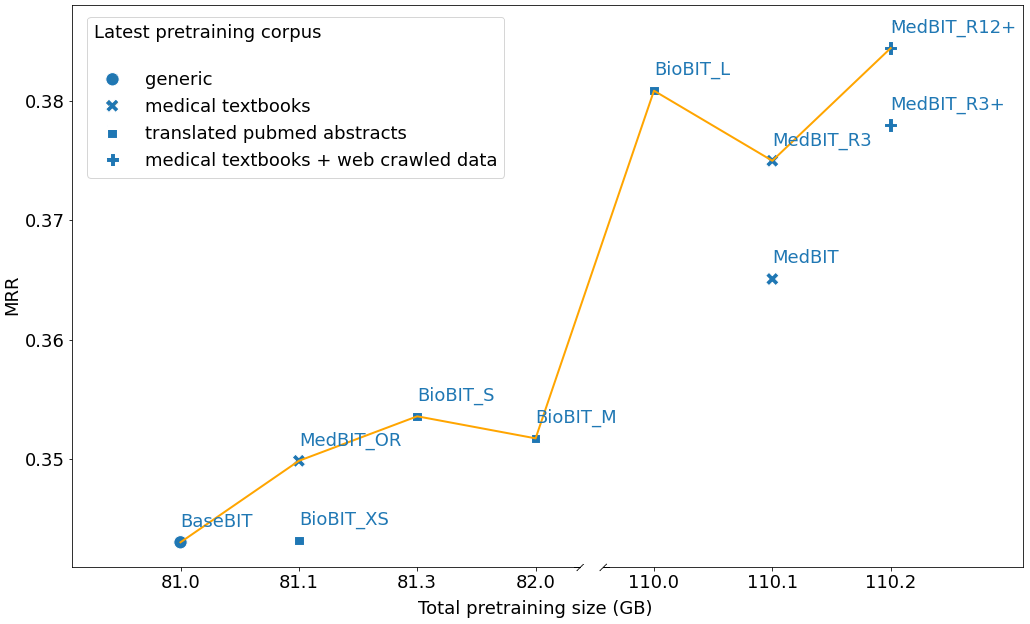}
\caption{Mean reciprocal rank progression over the different models while increasing the total pretraining size.}
\label{fig:trajectory}
\end{figure}

For what concerns the language modeling evaluation, results show a constant increment in the MRR as the corpus size increases, with the largest BioBIT\textsubscript{L} achieving a 14\% improvement on our test set compared with the baseline, as reported in Table 1. By looking at the MRR, it is also possible to monitor the emergence of biomedical knowledge as the corpus size grows, as shown in Table \ref{table:6}, where we can see how the term “memory” gradually raises in the ranked list of the model’s recommendations. A more extensive panoramic of the MRR progression is illustrated in Figure \ref{fig:trajectory}. 
On downstream tasks, the extent of improvement in F1 scores depends on the specific type of the task, with localized domain-specific models performing vastly better than the generic-purpose baselines on NER (worst case, +3\%) and RE (worst case, +6\%) but struggling in improving QA (+3\% on BioASQ 6b, no improvement on 4b and 5b), as reported in Table \ref{table:3}, Table \ref{table:5} and Table \ref{table:4} respectively. Variability between downstream tasks is expected, as the improvement depends not only on the complexity of the target task but also on the quality and size of the correspondent fine-tuning dataset, which is more than an order of magnitude smaller for QA, as reported in Table A1 in the Supplementary Notes.

\begingroup
\setlength{\tabcolsep}{0.5pt}
\begin{table}[]
\caption{Example of MRR calculation for Base and Bio BERT Italian models over a single sentence of the test corpus for MLM. The predicted token is extracted from the logit using a softmax and argmax transformation and reported in the table along with its probability.}
\label{table:6}
\begin{tabular}{@{}lccccc@{}}
\toprule
\multicolumn{6}{@{}l@{}}{\begin{tabular}[c]{@{}l@{}}\emph{IT: L'ipotesi più in voga è che nell’Alzheimer la regione dell’}\\\emph{ippocampo riduca la capacità di gestire la dopamina andan-}\\\emph{do a compromettere la [MASK] che è il principale sintomo}\\\emph{della patologia. }\\\\\emph{EN:  The most popular hypothesis is that in Alzheimer's}\\\emph{disease, the ability of the hippocampus to regulate dopamine}\\\emph{decreases, leading to [MASK] impairment which is the main}\\\emph{symptom of the disease.}\\\\ \emph{{[}MASK{]} = memoria / memory}\end{tabular}} \\ 
\midrule
Rank & BaseBIT & BioBIT\textsubscript{XS} & BioBIT\textsubscript{S} & BioBIT\textsubscript{M} & BioBIT\textsubscript{L} \\ \midrule
1\textsuperscript{st} & \begin{tabular}[c]{@{}c@{}}mobilità\\ 10\%\end{tabular} & \begin{tabular}[c]{@{}c@{}}funzione \\37\% \end{tabular}& \begin{tabular}[c]{@{}c@{}}depressione \\12\% \end{tabular}& \begin{tabular}[c]{@{}c@{}}\textbf{memoria} \\\textbf{34\%} \end{tabular}& \begin{tabular}[c]{@{}c@{}}\textbf{memoria} \\\textbf{53\%}\end{tabular} \\
2\textsuperscript{nd} & \begin{tabular}[c]{@{}c@{}}funzione \\6\% \end{tabular}& \begin{tabular}[c]{@{}c@{}}malattia \\6\% \end{tabular}& \begin{tabular}[c]{@{}c@{}}funzione \\11\% \end{tabular}& \begin{tabular}[c]{@{}c@{}}funzione \\33\% \end{tabular}& \begin{tabular}[c]{@{}c@{}}parola \\8\%\end{tabular} \\
3\textsuperscript{rd} & \begin{tabular}[c]{@{}c@{}}progressione \\4\% \end{tabular}& \begin{tabular}[c]{@{}c@{}}progressione \\5\% \end{tabular}& \begin{tabular}[c]{@{}c@{}}\textbf{memoria} \\\textbf{6\%} \end{tabular}& \begin{tabular}[c]{@{}c@{}}vista \\6\% \end{tabular}& \begin{tabular}[c]{@{}c@{}}vigilanza \\8\% \end{tabular}\\
4\textsuperscript{th} & \begin{tabular}[c]{@{}c@{}}patologia \\3\% \end{tabular}& \begin{tabular}[c]{@{}c@{}}\textbf{memoria} \\\textbf{4\%} \end{tabular}& \begin{tabular}[c]{@{}c@{}}patologia \\5\% \end{tabular}& \begin{tabular}[c]{@{}c@{}}percezione \\2\% \end{tabular}& \begin{tabular}[c]{@{}c@{}}funzione \\6\% \end{tabular}\\
5\textsuperscript{th} & \begin{tabular}[c]{@{}c@{}}\textbf{memoria} \\\textbf{3\%} \end{tabular}& \begin{tabular}[c]{@{}c@{}}depressione \\4\% \end{tabular}& \begin{tabular}[c]{@{}c@{}}malattia \\5\% \end{tabular}& \begin{tabular}[c]{@{}c@{}}visione \\2\% \end{tabular}& \begin{tabular}[c]{@{}c@{}}coscienza \\4\% \end{tabular}\\
\midrule
MRR & 0.20 & 0.25 & 0.33 & 1.00 & 1.00 \\ \bottomrule
\end{tabular}
\end{table}
\endgroup

On the other hand, our evaluation showed that the training on the Italian textbook corpus, limited in size but qualitatively superior, did not achieve significant improvements in upstream or downstream tasks by itself. The experiments conducted with MedBIT\textsubscript{OR} prove the data quantity constraint is indeed the prevalent factor limiting performance, even in the presence of qualitatively better datasets. However, the combination of the two strategies allowed us to succeed in achieving improved performance in several downstream tasks. The simple concatenation of the two pretraining iterations, though, was not sufficient to improve BioBIT\textsubscript{L}, appearing to be instead detrimental and leading to performance degradation both in MRR and F1 due to the catastrophic forgetting phenomenon, as shown in the MedBIT model. After the introduction of different CF mitigation techniques, the MedBIT\textsubscript{R3} model recovered from the performance degradation and managed to match the performances of the BioBIT checkpoint it was pretrained on, surpassing it when we augment our corpus with web-crawled data as in MedBIT\textsubscript{R3}\textsuperscript{+} and MedBIT\textsubscript{R12}\textsuperscript{+}. With the addition of only 0.7\% more, qualitatively higher, training data than the BioBIT checkpoint, the model achieves the best scores in NER (4 datasets out of 6) and RE (2 out of 2), thus proving the usefulness of local resources even when not abundant.

It is worth mentioning how the PPPL score appears to be significantly lower than BioBIT\textsubscript{L} for all the pretrained models based on the medical textbook corpus, even for the ones that do not perform well in terms of MRR. While not inherently connected with a downstream improvement, this intrinsic evaluation remains interesting, highlighting how the models pretrained on biomedical text originally written in Italian, like MedBIT\textsubscript{R12}\textsuperscript{+}, perceive input biomedical texts in Italian as more natural and syntactically correct than those pretrained on translated texts. On the other hand, our research indicates that measuring the MRR for MLM on a manually selected set of pertinent sentences and masked words can act as a solid indicator of the downstream behavior, allowing us to focus computational resources on the most promising biomedical models.

\section*{Conclusion}
Our study achieves its first objective of delivering an improved biomedical language model for Italian, reducing the gap with English. The BioBIT pretrained model is publicly available, serving as a starting point for Italian researchers and institutions interested in applying it to real-world setups. 
As stated in our second objective, we also provide a general workflow based on leveraging local resources and machine translation, applicable to different domain-specific scenarios and different languages as well. 
Finally, our findings on the effects of pretraining data size and quality highlight that quantity remains a rather rigid constraint for pretraining, despite the presence of qualitatively superior data tapped from local medical textbooks and specialized online assets, two sources that are commonly available also in non-English speaking countries. When minimal quantitative requirements are met, though, an additional pretraining round on such data can further push model performance. 
\subsection*{Limitations and Future Work}
Our investigation is limited by the amount of high-quality Italian sources we were able to collect in the scope of our study, a limitation we plan to overcome in follow-up research expanding our corpora with new resources. In future work, we also envision the collection of new downstream datasets based on original Italian biomedical text, as we believe the family of MedBIT models may have been penalized by being evaluated only on NMT-based biomedical datasets and not on text written natively in Italian. In particular, several fellow medical centers are currently working with us to extend our set and build a large, multicentric, database of real-world, annotated neuropsychiatric reports in Italian. Another limitation consists in testing only a single transformer-based architecture, i.e., BERT, over the plethora of variations produced by the prolific NLP community in the latest years. The computational cost of running the same battery of experiments for multiple architectures, however, would have been incompatible with the resources allocated for this work, therefore we decided to focus on the most researched one, as described in Methods.

\section*{Environmental Impact Statement}
The average computational cost we estimated for each pretraining run amounts to 3 GPU hours for the Italian textbook corpus and 720 GPU hours for the Pubmed corpus. Depending on the task and the size of the dataset, fine-tuning runs took 1 up to 3 GPU hours each, while automatic translation required 20 days to complete on Intel Xeon Gold 5218 CPUs. Experiments have been carried out on a Google Cloud virtual environment equipped with one Nvidia A100 40GB GPU, and on the IRCCS Centro San Giovanni di Dio Fatebenefratelli high-performance computing environment, equipped with four A100 GPUs. Based on local grid carbon intensities\footnotemark{}\footnotetext{https://cloud.google.com/sustainability/region-carbon} \footnotemark{}\footnotetext{https://www.isprambiente.gov.it/} and hardware power consumptions, the calculation described in Luccioni et al. \cite{luccioni_quantifying_2019} results in a total of approximately 273 kgCO\textsubscript{2}eq produced, which is equivalent to 1100 km driven by an average internal combustion engine car.

\section*{Conflict of interest statement}
None declared.

\section*{Acknowledgements}
The present work was partially funded by the National funding of the Italian Ministry of Health in the framework of the grant ISTITUTI NAZIONALI VIRTUALI (RCR 2020-23670067 and RCR‐2021‐23671214) and Ministry of Economy and Finance CCR-2017-23669078.

\printcredits

\bibliographystyle{plain}

\bibliography{biobit-refs}

\end{document}